# Adaptive Structure-constrained Robust Latent Low-Rank Coding for Image Recovery

Zhao Zhang[1,2,3], Lei Wang[1], Sheng Li[4], Yang Wang[2,3], Zheng Zhang[5], Zhengjun Zha[6] and Meng Wang[2,3]

[1] School of Computer Science and Technology, Soochow University, Suzhou 215006, China
[2] Key Laboratory of Knowledge Engineering with Big Data (Ministry of Education), Hefei University of Technology
[3] School of Computer Science and Information Engineering, Hefei University of Technology, Hefei, China
[4] Department of Computer Science, University of Georgia, 549 Boyd GSRC, Athens, GA 30602
[5] Bio-Computing Research Center, Harbin Institute of Technology (Shenzhen), Shenzhen, China
[6] School of Information Science and Technology, University of Science and Technology of China, Hefei, China
e-mail: cszzhang@gmail.com, layne.wright.wl@gmail.com

*Abstract*— In this paper, we propose a robust representation learning model called Adaptive Structure-constrained Low-Rank Coding (AS-LRC) for the latent representation of data. To recover the underlying subspaces more accurately, AS-LRC seamlessly integrates an adaptive weighting based block-diagonal structure-constrained low-rank representation and the group sparse salient feature extraction into a unified framework. Specifically, AS-LRC performs the latent decomposition of given data into a low-rank reconstruction by a block-diagonal codes matrix, a group sparse locality-adaptive salient feature part and a sparse error part. To enforce the block-diagonal structures adaptive to different real datasets for the low-rank recovery, AS-LRC clearly computes an auto-weighting matrix based on the locality-adaptive features and multiplies by the low-rank coefficients for direct minimization at the same time. This encourages the codes to be block-diagonal and can avoid the tricky issue of choosing optimal neighborhood size or kernel width for the weight assignment, suffered in most local geometrical structures-preserving low-rank coding methods. In addition, our AS-LRC selects the $L_{2,1}$-norm on the projection for extracting group sparse features rather than learning low-rank features by Nuclear-norm regularization, which can make learnt features robust to noise and outliers in samples, and can also make the feature coding process efficient. Extensive visualizations and numerical results demonstrate the effectiveness of our AS-LRC for image representation and recovery.

## I. INTRODUCTION

WITH the continuous development of Internet, transmitted information or data (e.g., image) in daily communication are getting more and more complicated due to presence of redundant information, high-dimensionality, corruptions and noise. This potentially makes the task of learning the underlying subspaces from real images still challenging. Since image data can usually be characterized by low-rank structures, robust low-rank subspace recovery and representation by minimizing the Nuclear-norm based formulation has arousing much attention in the recent years [1-11][26-28][31-34][37-43].

Two most representative low-rank representation models are *Robust Principal Component Analysis* (RPCA) [1-2] and *Low-Rank Representation* (LRR) [3]. Both RPCA and LRR aim at decomposing given data matrix into a low-rank component and a sparse error, where the low-rank component corresponds to the recovered representations. By low-rank coding, RPCA and LRR are able to handle the corruptions and correct errors jointly. RPCA assumes that data are from a single low-rank subspace, while LRR considers a more general case that the samples are approximately drawn from a union of low-rank subspaces. Thus, LRR can handle mixed data, which is an attractive property. But note that LRR can only work well under the assumption that all subspaces are independent, while in fact this assumption cannot be guaranteed in certain real-world applications [4]. To address this problem, *Structure-Constrained Low-Rank Representation* (SC-LRR) [4], as an extension of LRR, was recently proposed to analyze the multiple linear disjoint subspace structures with a pre-defined affinity matrix, which can also provide a more general view for handling real vision data. Most existing LRR methods aim to approximate the block-diagonal representation matrix by using different structure priors. By directly designing a soft block diagonal regularizer, which encourages a matrix to be or close to k-block diagonal, *Block Diagonal Representation* (BDR) [5] recovers the subspaces by using the block diagonal structure prior of the representation in the ideal case.

Although aforementioned RPCA, LRR, SC-LRR and BDR methods can recover the underlying low-rank subspace(s) to some extent, they are essentially transductive algorithms, i.e., they all cannot handle new data efficiently. Specifically, given a new sample, RPCA, LRR and SC-LRR have to recalculate over all the data again leading to high computational cost and making them inept to the real applications that need fast online computation. Towards handling this issue, some joint low-rank recovery and salient feature extraction based representation frameworks were presented, e.g., *Inductive Robust Principal Component Analysis* (IRPCA) [6], *Latent LRR* (LatLRR) [7], *Frobenius norm based Latent LRR* (FLLRR) [8] and *Inductive Low-rank and Sparse Principal Feature Coding* (I-LSPFC) [9]. IRPCA calculates a low-rank projection jointly to project the samples into their respective subspaces, and the projection can also involve new data efficiently. LatLRR improves LRR using unobserved hidden data to extend the dictionary to overcome the insufficient data sampling issue, but LatLRR tends to have high computational cost for the optimization of double nuclear norm. FLLRR approximates the rank function by employing the Frobenius norm to replace the nuclear norm for efficiency. I-LSPFC incorporates embedded low-rank and sparse features by a projection into one problem for direct minimization.

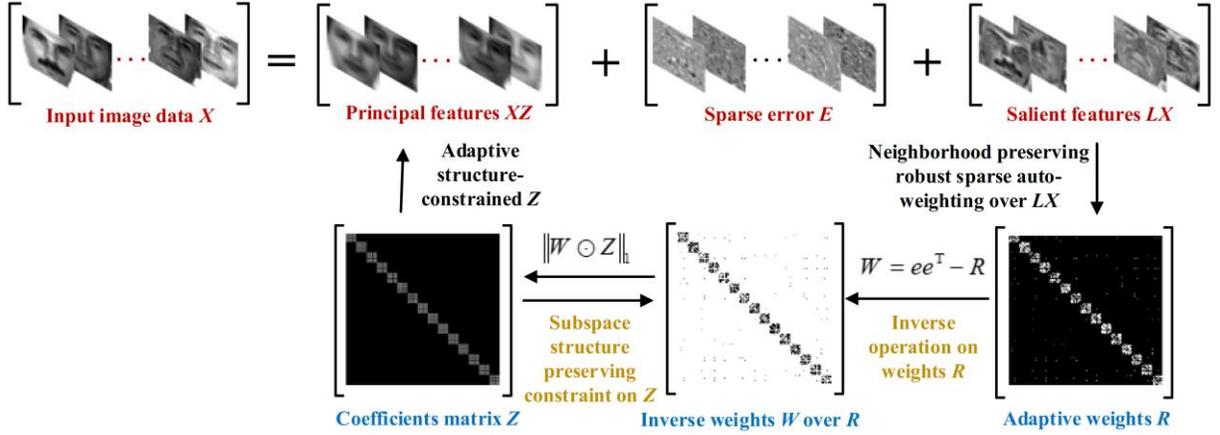

Fig. 1: The flow-diagram of our proposed AS-LRC algorithm for robust low-rank recovery and salient feature extraction.

It is worth noting that although IRPCA, LatLRR, FLLRR and I-LSPFC handle the out-of-sample issue, they cannot preserve of local manifold structures and similarity among the samples, which may result in the inaccurate representation. To obtain the locality preserving representation, certain manifold based low-rank coding models were recently presented, such as *Laplacian Regularized LRR* (rLRR) [10], *Similarity-Adaptive Latent LRR* (SA-LatLRR) [11] and *Non-negative Sparse Hyper-Laplacian regularized LRR* (NSHLRR) [12]. rLRR is able to preserve local information by pre-calculating a graph weight matrix using the Gaussian function and defining a Laplacian matrix to preserve the locality, similarly as SC-LRR. Thus, rLRR and SC-LRR will suffer from the tricky issue of choosing the optimal kernel width and neighborhood size for assigning weights. NSHLRR also employs a hypergraph Laplacian regularizer to pre-obtain a so-called hyper-Laplacian to represent low-dimensional structures and intrinsic geometry in data. But note that the pre-definition manner of rLRR, SC-LRR and NSHLRR cannot ensure the pre-calculated Laplacian to be optimal for the subsequent low-rank coding. In addition, the weighting and Laplacian in rLRR, SC-LRR and NSHLRR are computed in the original input space that usually contains various noise and errors, therefore the resulted similarities may be inaccurate for locality preservation. Hence, it will be better to present a joint and robust weighting learning strategy to constrain the structures of coefficients and preserve the locality of samples in an adaptive manner.

In this paper, we propose a novel robust adaptive learning framework to solve the drawbacks mentioned above. The main contributions of this paper are shown as follows:

(1) Technically, a Robust Adaptive Structure-constrained Low-Rank Coding algorithm termed AS-LRC is proposed for the latent representation. AS-LRC is motivated by the ideas of LatLRR and SC-LRR, so it owns all merits of both LatLRR and SC-LRR for joint salient feature extraction, constraining and analyzing the multiple linear disjoint subspace structures. But more importantly, our AS-LRC overcomes their shortcomings. Specifically, AS-LRC performs latent decomposition of given data into a low-rank reconstruction by a block-diagonal codes matrix, a group sparse locality-adaptive salient feature part and a sparse error part. To recover multiple subspaces accurately, AS-LRC seamlessly integrates the adaptive weighting guided block-diagonal structure-constrained low-rank representation and group sparse salient feature extraction into a unified model. Thus, AS-LRC will be a more general framework for low-rank multiple subspaces recovery and subspace learning.

(2) To make the block-diagonal structures of the coefficients adaptive to different real datasets for low-rank recovery, AS-LRC computes an auto-weighting matrix based on the locality-adaptive salient sparse features rather than the original data, and then multiplies by the low-rank coefficients for minimization. This operation encourages the coefficients to be block-diagonal and can also avoid the tricky issue of choosing optimal neighbor number and kernel width for the weight assignment, suffered in rLRR and SC-LRR. Besides, sparse $L_{2,1}$-norm is also imposed on the adjacency matrix to ensure the sparse properties of learnt weights. By the joint weight learning, AS-LRC can explicitly ensure the learned weights to be joint-optimal for representation and the subsequent low-rank coding, which clearly differs from SC-LRR. Due to the adaptive weighting, local information of salient features can be clearly preserved in the coding space.

(3) For extracting salient features efficiently, AS-LRC uses the $L_{2,1}$-norm on the projection for extracting the group sparse features rather than learning low-rank features by Nuclear-norm regularization, which can also make the feature extraction step robust to noise and outliers in given data.

The outline of this paper is shown as follows. In Section II, we briefly describe the related work. In Section III, we present the formulation, optimization and classification strategy of our AS-LRC. Section IV shows the simulation settings and results. Section V draws the conclusion of this paper.

## II. RELATED WORK

In this section, we introduce the related work that are closely related to our proposed algorithm.

### A. Latent Low-Rank Representation (LatLRR)

Given a set of samples $X = [x_1,...,x_N] \in \mathbb{R}^{d \times N}$, where $d$ denotes the original dimension and $N$ is the number of samples, LatLRR mainly improves LRR by using the unobserved hidden data to extend the dictionary and thus overcome the insufficient data sampling issue. For recovering hidden data, LatLRR considers the following Nuclear-norm minimization problem:

$$\min_Z \|Z\|_*, \ s.t. \ X_O = [X_O, X_H]Z, \quad (1)$$

where $\|Z\|_*$ is the nuclear-norm of $Z$ [3-5], i.e., the sum of its singular values, $X_O$ is the observed data and $X_H$ is unobserved hidden data. Suppose $Z_{O,H} = [Z_{O|H}, Z_{H|O}]$ is the solution to this problem, where $Z_{O|H}$ and $Z_{H|O}$ respectively correspond to $X_O$ and $X_H$, and let $U\Sigma V$ denote the skinny SVD of $[X_O, X_H]$ and partition matrix $V$ as $V = [V_O, V_H]$ such that $X_O = U\Sigma V_O^T$ and $X_H = U\Sigma V_H^T$, the above constraint can be simplified as

$$\begin{aligned} X_O &= X_O Z_{O|H} + X_H Z_{H|O} \\ &= X_O Z_{O|H} + X_H V_H V_O^T \\ &= X_O Z_{O|H} + U\Sigma V_H^T V_H V_O^T \\ &= X_O Z_{O|H} + L_{H|O} X_O \end{aligned} \quad (2)$$

where $L_{H|O} = U\Sigma V_H^T V_H \Sigma^{-1} U^T$. Since $Z_{O|H}$ and $L_{H|O}$ should be of low-rank, one can let $Z$ and $L$ denote $Z_{O|H}$ and $L_{H|O}$ simply. As the L1-norm is applied on $E$, LatLRR recovers the hidden effects by minimizing the following problem:

$$\min_{Z,L,E} \|Z\|_* + \|L\|_* + \lambda \|E\|_1, \ s.t. \ X = XZ + LX + E, \quad (3)$$

where $XZ$, $LX$ and $E$ denote principal features, salient features and sparse errors respectively, and $\lambda$ is a positive parameter depending on the level of noise [1-3]. Note that no adjustable parameter is used between $Z$ and $L$ since they can be balanced automatically according to the conclusion of [7].

*B. Structure-Constrained LRR (SC-LRR)*

SC-LRR is an extension of original LRR by computing a weight matrix to leverage the structure of multiple disjoint subspaces to make the representation accurate, which is more general for real-world vision data. Because it is known that the standard LRR algorithm can work well under the assumption that all subspaces are independent, but note that this assumption cannot be ensured in certain real-world problems. The objective function of SC-LRR can be defined as the following problem:

$$\min_Z \|Z\|_* + \beta \|W \odot Z\|_1 + \lambda \|E\|_{2,1} \ s.t. \ X = XZ + E, \quad (4)$$

where $W$ denotes a predefined weight matrix and $\odot$ denotes the Hadamard product between two matrices [4]. The Hadamard product between $W$ and $Z$ can be defined as

$$W \odot Z = \begin{bmatrix} W_{11}Z_{11} & W_{12}Z_{12} & \dots & W_{1n}Z_{1n} \\ W_{21}Z_{21} & W_{22}Z_{22} & \dots & W_{2n}Z_{2n} \\ \vdots & \vdots & \ddots & \vdots \\ W_{m1}Z_{m1} & W_{m2}Z_{m2} & \dots & W_{mn}Z_{mn} \end{bmatrix}. \quad (5)$$

The minimization of the Hadamard product $W \odot Z$ can enable SC-LRR to leverage the structure of multiple disjoint subspaces. By the observation that if the angle between data points is small they are likely to be in the same class, SC-LRR normalizes the data and compute the absolute value of the inner product. In this way, the weight determined by angle can be defined as

$$W_{i,j} = 1 - exp\left(-\frac{1 - |x_i^{*T} x_j^*|}{\sigma}\right), \quad (6)$$

where $x_i^*$ and $x_j^*$ denote the normalized samples of $x_i$ and $x_j$ respectively, and $\sigma$ is empirically set as the mean of elements of $B$ where $B_{i,j} = 1 - |x_i^{*T} x_j^*|$. In general, this definition induces a smaller value for weighting the data points of the same class, but a larger weight between data points from different classes.

### III. ADAPTIVE STRUCTURE-CONSTRAINED LOW-RANK CODING FOR IMAGE RECOVERY

*A. Proposed Formulation*

In this section, we formulate the objective function of AS-LRC. Our model improves the representation power by analyzing the multiple disjoint subspace structures in an adaptive manner, and extends for joint extraction of group sparse salient features $LX$, where $L$ is the projection for feature extraction. To make the subspace structures adaptive to various datasets for the latent subspace recovery, AS-LRC jointly obtains an auto-weighting matrix $R$ by minimizing the reconstruction error $\|LX - LXR\|_F^2$ based on the locality-adaptive salient features $LX$:

$$\wp(L,R) = \|LX - LXR\|_F^2 + \|e^T - e^T R\|_F^2 + \|R\|_{2,1}, \quad (7)$$

where the sum-to-one constraint $\|e^T - e^T R\|_F^2$ is to force the sum of each column in the adaptive weight matrix $R$ to be close to 1, which can preserve the locality of the adaptive weights to some extent. The L$_{2,1}$-norm based $R$ can make many rows of $R$ to be zeros, i.e., group sparsity. As a result, it can select the most important samples to reconstruct given data for more accurate reconstruction and similarity measure. As a result, the jointly obtained reconstruction weights will be good for the subsequent subspace recovery, and can also avoid the difficult problem of choosing optimal neighborhood size or kernel width. Besides, the auto-weighting procedures can preserve local neighborhood information of samples in an adaptive manner.

To force the representation coefficients $Z$ have block-diagonal structures, AS-LRC clearly regularizes the adaptive weighting matrix $R$ guided structure-constraint on $Z$, i.e., $\|(ee^T - R) \odot Z\|_1$. Where $ee^T$ denotes a matrix of all ones so that this structure-preserving constraint can induce a smaller value for weighting the data points in the same subspace, but larger values to weight the data points from different subspaces, similarly as [4]. This is also another reason for normalizing $Z$. Therefore, AS-LRC can handle the recovery issue of disjoint subspaces effectively. Thus, the problem of AS-LRC can be formulated as

$$\min_{Z,L,E,R} \|Z\|_* + \|L\|_{2,1} + \alpha \|(ee^T - R) \odot Z\|_1 + \beta \wp(L,R) + \lambda \|E\|_1, \quad (8)$$
$$s.t. \ X = XZ + LX + E$$

where the L$_{2,1}$-norm is regularized on the projection $L$, which can make the feature embedding learning robust to noise and outliers in data when delivering group sparse features, and can also make the feature learning process efficient.

To simplify the expression, we involve an auxiliary variable $W$ to represent $ee^T - R$. By substituting $\wp(R)$ back into Eq.(8), the final objective function of AS-LRC is obtained as

$$\min_{Z,L,E,W,R} \|Z\|_* + \|L\|_{2,1} + \alpha \|W \odot Z\|_1 + \beta \left(\|A - AR\|_F^2 + \|R\|_{2,1}\right) + \lambda \|E\|_1$$
$$s.t. \ X = XZ + LX + E, \ W = ee^T - R, \ A = \left(X^T L^T, e\right)^T$$
$$(9)$$

The flow-diagram of our AS-LRC algorithm is shown in Fig. 1. It is clear that our AS-LRC unifies the procedures of the auto-weighting based structure-preserving constraint and the latent subspace representation. That is, our AS-LRC can preserve the neighborhood information in an adaptive manner with low rank coding. Next, we describe the optimization procedures.

*B. Optimization*

The objective function of our AS-LRC is generally convex, so it can be solved by various methods. In this study, we use the Inexact Augmented Lagrange Multiplier (Inexact ALM) [13] method for efficiency. Following the common procedures [7][9], we introduce some auxiliary variables $J$, $F$, $Q$ and $S$ to make the problem easily solvable. We first convert Eq.(9) into

$$\min_{\substack{Z,L,E,W,R \\ J,F,S,Q}} \|J\|_* + \|F\|_{2,1} + \alpha \|W \odot Q\|_1 + \beta \left(\|A - AR\|_F^2 + \|S\|_{2,1}\right) + \lambda \|E\|_1,$$

$$s.t.\ X = XZ + LX + E, Z = J, L = F, Z = Q, R = S, W = ee^T - R$$
(10)

where $A = \left(X^T L^T, e\right)^T$. The augmented Lagrangian function $\ell$ of the above problem can be defined as

$$\ell = \|J\|_* + \|F\|_{2,1} + \alpha \|W \odot Q\|_1 + \beta \left(\|A - AR\|_F^2 + \|S\|_{2,1}\right) + \lambda \|E\|_1$$
$$+ \langle Y_1, X - XZ - LX - E \rangle + \langle Y_2, Z - J \rangle + \langle Y_3, L - F \rangle + \langle Y_4, Z - Q \rangle$$
$$+ \langle Y_5, R - S \rangle + \langle Y_6, ee^T - W - R \rangle + \frac{\mu}{2} \left( \|X - XZ - LX - E\|_F^2 + \right.$$
$$\left. \|Z - J\|_F^2 + \|L - F\|_F^2 + \|Z - Q\|_F^2 + \|R - S\|_F^2 + \|ee^T - R - W\|_F^2 \right)$$
(11)

where $Y_1$, $Y_2$, $Y_3$, $Y_4$, $Y_5$ and $Y_6$ are Lagrangian multipliers, $\mu$ is a weighting factor and $\langle A, B \rangle = tr\langle A^T B \rangle$ is the inner-product. As the involved variables depend on each other, they cannot be solved directly, so we use an alternative updating strategy to update one of them by fixing others at each time, i.e.,

*1) Fix others, update L, Z and R:*

We first fix the weight matrix $R$ and coefficients $Z$ to update the projection $L$. By removing terms that are independent of $L$ from Eq. (11), taking the derivative $\partial\ell/\partial L$ w.r.t. $L$ and setting it to zero, we can easily update the projection $L$ as

$$L_{k+1} = \left(Y_1^k X^T - Y_3^k + \mu_k \left(X - XZ_k - E_k\right) X^T + \mu_k F_k\right) \times$$
$$\left(2\beta \left(X - XR_k\right)\left(X - XR_k\right)^T + \mu_k \left(XX^T + I\right)\right)^{-1}.$$
(12)

Similarly, by taking the derivative $\partial\ell/\partial Z$ and setting it to 0, we can update the coding coefficient matrix $Z$ as

$$\Xi_{k+1} = \left(X^T Y_1^k - Y_2^k - Y_4\right)/\mu_k + X^T \left(X - L_{k+1} X - E_k\right) + J_k + Q_k$$
$$Z_{k+1} = \left(2I + X^T X\right)^{-1} \Xi_{k+1}$$
(13)

By taking the derivative $\partial\ell/\partial R$, we can update weights $R$ as

$$\Theta_{k+1} = Y_6^k - Y_5^k - 2\beta \left(L_{k+1} X\right)^T L_{k+1} X - 2\beta ee^T + \mu_k \left(ee^T - W_k\right) + \mu S_k$$
$$R_{k+1} = \left(\mu_k I - 2\beta \left(L_{k+1} X\right)^T L_{k+1} X - 2\beta ee^T\right)^{-1} \Theta_{k+1}.$$
(14)

*2) Fix others, update Q and W:*

In this step, we focus on updating $Q$ and $W$ by fixing the other variables. By removing the irrelevant items from the augmented Lagrangian function, the matrix $Q$ can be updated by

$$Q_{k+1} = \arg\min_Q \frac{\alpha}{\mu_k} \|W_k \odot Q\|_1 + \frac{1}{2} \left\| Q - \left(Z_{k+1} + Y_4^k/\mu_k\right) \right\|_F^2.$$
(15)

Let $\Psi^Q = Z_{k+1} + Y_4^k/\mu_k$, and let $\Psi^Q_{i,j}$ be the $i$-th row and the $j$-th column of matrix $\Psi^Q$. The $(i, j)$-th element $Q_{k+1}^{i,j}$ of $Q_{k+1}$ at the $(k+1)$-th iteration is updated as $Q_{k+1}^{i,j} = \Delta_{\varepsilon_{i,j}}\left[\Psi^Q_{i,j}\right]$ by the scalar shrinkage operator, where $\varepsilon_{i,j} = (\alpha/\mu_k) W_{i,j}^k$. Similar to the optimization of $Q$, we can update $W$ by

$$W_{k+1} = \arg\min_W \frac{\alpha}{\mu_k} \|W \odot Q_{k+1}\|_1 + \frac{1}{2} \left\| W - \left(ee^T - R_{k+1} + Y_6^k/\mu_k\right) \right\|_F^2.$$
(16)

*3) Fix others, update variables J, F, S and error E:*

For the optimization of $J$, other variables are fixed as constants. By removing the terms irrelevant to $J$ and using the Singular Value Thresholding (SVT) operator [25], we ca update $J$ as

$$J_{k+1} = \arg\min_J \frac{1}{\mu_k} \|J\|_* + \frac{1}{2} \left\| J - \left(Z_{k+1} + Y_2^k/\mu_k\right) \right\|_F^2,$$
$$= \Omega_{1/\mu_k} \left[Z_{k+1} + Y_2^k/\mu_k\right]$$
(17)

where $\Omega_{1/\mu_k}\left[Z_{k+1} + Y_2^k/\mu_k\right] = U\Delta_{1/\mu_k}[\Sigma]V$ is the singular value shrinkage operator [7], $U\Sigma V$ is the SVD of $Z_{k+1} + Y_2^k/\mu_k$, and $\Delta_\varepsilon [x] = \text{sgn}(x)\max(|x| - \varepsilon, 0)$ is the scalar shrinkage operator. After $L_{k+1}$ is computed, the solution of the sparse error matrix $F$ can be analogously inferred as

$$F_{k+1} = \arg\min_F \frac{1}{\mu_k} \|F\|_{2,1} + \frac{1}{2} \left\| F - \left(L_{k+1} + Y_3^k/\mu_k\right) \right\|_F^2.$$
(18)

Let $\Psi^F = L_{k+1} + Y_3^k/\mu_k$ and let $\Psi^F_{:,i}$ denote the $i$-th column of matrix $\Psi^F$. The $i$-th column $F_{k+1}^{:,i}$ of solution $F_{k+1}$ at the $(k+1)$-th iteration can be defined as [3]:

$$F_{k+1}^{:,i} = \begin{cases} \frac{\|\Psi^F_{:,i}\|_2 - 1/\mu_k}{\|\Psi^F_{:,i}\|_2} \Psi^F_{:,i}, & \text{if } \|\Psi^F_{:,i}\|_2 > 1/\mu_k \\ 0, & \text{otherwise} \end{cases}.$$
(19)

Similar to the optimization of variable $F$, the solution of the auxiliary variable $S$ can be analogously inferred as

$$S_{k+1} = \arg\min_S \frac{\beta}{\mu_k} \|S\|_{2,1} + \frac{1}{2} \left\| S - \left(R_{k+1} + Y_5^k/\mu_k\right) \right\|_F^2.$$
(20)

By removing terms irrelevant to $E$, we can update the sparse error $E$ using the scalar shrinkage operator as

$$E_{k+1} = \arg\min_E \frac{\lambda}{\mu_k} \|E\|_1 + \frac{1}{2} \left\| E - \left(X - XZ_{k+1} - L_{k+1}X + Y_1^k/\mu_k\right) \right\|_F^2$$
$$= \Delta_{\lambda/\mu_k}\left[X - XZ_{k+1} - L_{k+1}X + Y_1^k/\mu_k\right]$$
(21)

For complete presentation of our method, we summarize the optimization procedures in the Algorithm 1. Note that we find experimentally that our model can generally perform well and converge with the number of iterations ranging from 50 to 150 in most cases. Note that the major computational cost of the Algorithm 1 is computing the SVD of matrix in Step 1. Thus, the time complexity of our approach is the same as the existing LRR and IRPCA, etc. Hence, the computation is also efficient, especially when $n$ is a relatively small, similarly as IRPCA and

**Algorithm 1: Our AS-LRC solved by Inexact ALM**

**Inputs**: Training data matrix $X = [x_1, x_2, ..., x_N] \in \mathbb{R}^{d \times N}$, tuning parameters $\alpha$, $\beta$ and $\lambda$;
**Initialization**: $E_0 = 0$, $J_0 = Z_0 = Q_0 = 0$, $R_0 = S_0 = W_0 = 0$, $L_0 = F_0 = 0$, $Y_1^0 = 0$, $Y_2^0 = 0$, $Y_3^0 = 0$, $Y_4^0 = 0$, $Y_5^0 = 0$, $Y_6^0 = 0$, $\max_\mu = 10^{10}$, $\mu_k = 10^{-6}$, $\eta = 1.12$, $\varepsilon = 10^{-6}$, $k = 0$;
**While not converged do**
1. Fix others, update low-rank codes $Z_{k+1}$ by Eq.(13);
2. Fix others, update sparse projection $L_{k+1}$ by Eq.(12);
3. Fix others, update the sparse error $E_{k+1}$ by Eq.(21);
4. Fix others, update the weight matrix $R_{k+1}$ by Eq.(14);
5. Fix others, update the auxiliary variables $J_{k+1}$, $F_{k+1}$, $Q_{k+1}$, $W_{k+1}$ and $S_{k+1}$ by Eqs. (17), (18), (15), (16) and (20);
6. Update the Lagrange multipliers:
   $Y_1^{k+1} = Y_1^k + \mu_k(X - XZ_{k+1} - L_{k+1}X - E_{k+1})$,
   $Y_2^{k+1} = Y_2^k + \mu_k(Z_{k+1} - J_{k+1})$, $Y_3^{k+1} = Y_3^k + \mu_k(L_{k+1} - F_{k+1})$,
   $Y_4^{k+1} = Y_4^k + \mu_k(Z_{k+1} - Q_{k+1})$, $Y_5^{k+1} = Y_5^k + \mu_k(R_{k+1} - S_{k+1})$,
   $Y_6^{k+1} = Y_6^k + \mu_k(ee^\top - W_{k+1} - R_{k+1})$, ;
7. Update the parameter $\mu$ with $\mu_{k+1} = \min(\eta\mu_k, \max_\mu)$;
8. Convergence check: if $\max(\|X - XZ_{k+1} - L_{k+1}X - E_{k+1}\|_\infty$, $\|Z_{k+1} - J_{k+1}\|_\infty, \|L_{k+1} - F_{k+1}\|_\infty, \|Z_{k+1} - Q_{k+1}\|_\infty, \|R_{k+1} - S_{k+1}\|_\infty$, $\|W_{k+1} - ee^\top - R_{k+1}\|_\infty) < \varepsilon$, stop; else $k=k+1$.

**End while**
**Outputs**: $Z^* = Z_{k+1}$, $L^* = L_{k+1}$, $E^* = E_{k+1}$.

LRR. About the convergence of our method, we must say that the convergence property of inexact ALM has been well studied when the number of blocks is two and it can generally perform well in reality [1-3]. But it is still tricky to theoretically prove the convergence of the method containing over two blocks.

### C. Approach for Out-of-sample Classification

We discuss the method of using AS-LRC for including outside new data. Specifically, we present an embedding based label prediction approach by using the computed locality-preserving salient features $L^*X$ of AS-LRC. Based on salient features $L^*X$, we further train a linear classifier $C$ when the labels of training samples are known. Denote by $H = [l(x_1), l(x_2), ..., l(x_N)] \in \mathbb{R}^{c \times N}$ the class label set, where $c$ is the number of classes, $N$ is the number of samples, and $l(x_i)$ indicates the label of $x_i$. Note that each $l(x_i)$ has only one nonzero entry and the position of the one nonzero entry in $l(x_i)$ determines the label of $x_i$. We define the following problem for learning a linear classifier:

$$\langle C, E^C \rangle = \arg\min_{C, E^C} \|E^C\|_1, \; s.t. \; H^\top = (L^*X)^\top C + E^C, \quad (22)$$

where $E^C$ is classification error. Based on simple computation, one can easily obtain a classifier $C^*$. Then, we are ready to show how to classify the new data efficiently. Specifically, given a new test sample $x_{new}$, one can obtain its class label as

$$l(x_{new}) = \arg\min_i \left(s_{new} = C^{*\top} L^* x_{new}\right)_i, \quad (23)$$

i.e., embedding sample $x_{new}$ onto $L^*$ and $C^{*\top}$ sequentially, where $s_{new} = C^{*\top} L^* x_{new}$ is the soft label vector of $x_{new}$ and the final hard label can be obtained from the codes $s_{new}$. That is, the position corresponding to the biggest value in $s_{new}$ decides the label of $x_{new}$. In reality, we can handle entire test set $X_{test}$ straightly to obtain a whole class label matrix $S_{test} = C^{*\top} L^* X_{test}$. Thus, our AS-LRC algorithm can handle the label prediction task efficiently.

## IV. EXPERIMENTAL RESULTS AND ANALYSIS

In this section, we perform visual and numerical experiments to illustrate the effectiveness of our AS-LRC method, and show the comparison results with closely related low-rank or sparse coding models, including PCA, PCA-L1, IRPCA, LatLRR, SA-LatLRR, FLLRR, rLRR and I-LSPFC. After the projection of each algorithm is achieved, we learn a classifier $C^*$ according to Eq. (22) for fair comparison. Since each low-rank or sparse coding algorithm has a common parameter $\lambda$, it is carefully chosen for each method. The other parameters of each model are also carefully chosen for the fair comparison. There are two model parameters $\alpha$ and $\beta$ in our AS-LRC, which are selected from the candidate set $\{10^{-8}, 10^{-6}, ..., 10^6, 10^8\}$, and the best results over tuned parameters are reported for evaluation. In this study, seven face databases, one handwriting database and three object databases are evaluated. The detailed information of used real-world databases is described in Table 1. By following the common practice, all the images are down-sampled into $32 \times 32$ pixels, thus each image corresponds to a data point in a 1024-dimensional space. All the experiments are carried out on a PC with Intel(R) Core(TM) i5-4590 @ 3.30Hz 8.00GB.

**Table 1:** List of used datasets and detailed information

| Dataset Name | Data Type | Classes($c$) | Points($N$) |
|---|---|---|---|
| UNIPEN | Handwriting | 93 | 62382 |
| MIT CBCL | Face image | 10 | 3240 |
| ORL | Face image | 40 | 400 |
| Yale | Face image | 15 | 165 |
| YaleB-UMIST | Face image | 58 | 3426 |
| JAFFE | Face image | 10 | 213 |
| AR | Face image | 100 | 2600 |
| CMU PIE | Face image | 68 | 11554 |
| COIL-20 | Object image | 20 | 1440 |
| Caltech 101 | Object image | 101 | 9111 |
| ETH80 | Object image | 80 | 3280 |

### A. Visual Image Analysis by Data Recovery

We first evaluate AS-LRC for representing handwriting images and face images, and we mainly compare the recovery results with those of LatLRR, SA-LatLRR and FLLRR. For a given data matrix $X$, each method decomposes it to a low-rank part (i.e., $L^*$ for RPCA, $P^*X$ for IRPCA, $XZ^*$ for LatLRR, and $XZ^*$ for our algorithm) and a sparse error part $E^*$, where the low-rank component denotes the recovered result of each algorithm.

**Handwriting Representation.** We first test the performance of each method for representing the handwriting images by visualizing the low-rank representation of images. In this study, the UNIPEN handwriting dataset [16] which contains 93 categories and a total of 62,382 image characters in grayscale background is used. The characters in UNIPEN database are assigned using the ASCII numeric code from 33 to 126. The original images, recovered images, salient features, and error images are exhibited in Fig. 2. We can easily find that AS-LRC can not only recover the missing or irregular handwriting strokes accurately than its competitors, but also preserve the important features. Note that we have used red and green solid rectangles to highlight the advantages of AS-LRC over other algorithms.

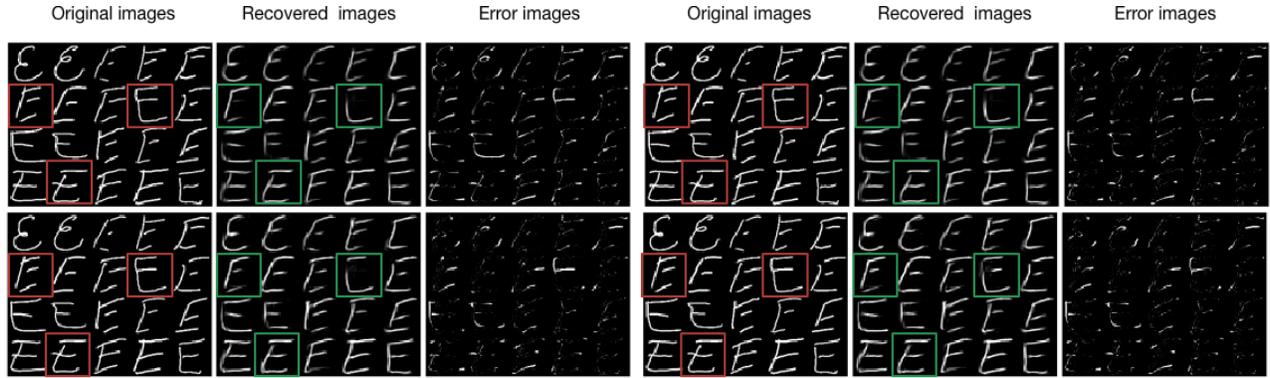

**Fig. 2**: Handwriting recovery results of LatLRR (top left), FLLRR (bottom left), SA-LatLRR (top right) and our AS-LRC (bottom right) based on the UNIPEN database.

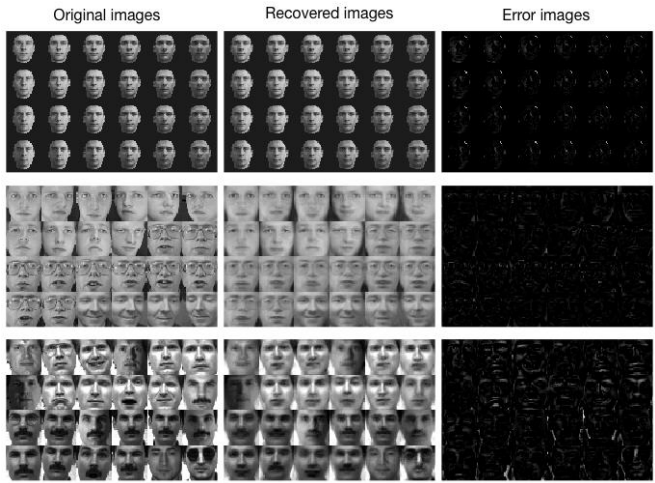

**Fig. 3**: Face recovery results of AS-LRC on MIT CBCL (first row), ORL (second row) and Yale datasets (third row).

*Face Representation.* We evaluate each model for face image representation. Three face databases, i.e., MIT CBCL (http://cbcl.mit.edu/software-datasets/heisele/facerecognitiondatabase.html), ORL [17] and Yale [18] are involved. In this study, we use MIT CBCL to test our method to recover rotating face poses, use ORL for recovering facial expressions, and use Yale to recover to facial expressions and occlusions by glasses. For each database, we choose 24 images to form a data matrix X of size $128 \times 192$. The recovered images are shown in Fig. 3. We see that AS-LRC can recover the face images of different poses, expressions and occlusions effectively.

### B. Quantitative Evaluation of Image De-noising

We quantify the de-nosing results of each algorithm, including LRR, LatLRR, BDR, NSHLRR, SA-LatLRR, SC-LRR, rLRR, FLLRR and AS-LRC. The reconstruction accuracy $\zeta_{acc}$ [9] is used as the quantitative evaluation metric for image de-nosing by reconstruction and embedding.

*Recovering objects with pixel corruptions.* COIL-20 database (www.cs.columbia.edu/CAVE/software/softlib/coil-20.php) is used in this study. We add white Gaussian noise with SNR=10 to object images and corrupt a percentage of randomly selected pixels from the images by replacing the original values with corrupted values. The corrupted pixels are randomly selected. In this study, we choose the car object images to create a data matrix of dimension $64 \times 96$, and we vary the percentage of the corrupted pixels from 0% to 90%. We illustrate the averaged quantitative de-nosing result of each method over 15 randomly selected pixels to be corrupted in Fig. 4. The reconstruction accuracies based on *Z* and *L* are respectively shown in Fig. 4(a) and Fig. 4(c). Note that LatLRR, SA-LatLRR, rLRR, FLLRR and AS-LRC can deliver a projection for embedding samples, thus we only describe their de-nosing results by embedding in Fig. 4(c). We find that our AS-LRC outperforms other methods by delivering higher accuracies, especially for the de-nosing results by embedding with *L*, and the performance difference among the tested methods becomes larger as the percentage of corruptions is increased. That is, our AS-LRC can recover the corruptions more accurately than the compared methods.

*Recovering faces with pixel corruptions.* JAFFE face dataset [19] is evaluated. Each face image has been rated on 6 emotion adjectives by 60 Japanese females. In this study, we corrupt a percentage of randomly selected pixels from the face images by replacing the original gray value with inverted values, i.e., the corrupted value g is replaced by 256-g. We vary the percentage of corrupted pixels from 0% to 90%. The averaged de-nosing results are shown in Fig. 5. We can find that our method still delivers higher accuracy than other competitors in most cases.

### C. Image Recognition vs. Different Training Numbers

We examine each method for image recognition by varying the numbers of training samples. One face and one object databases are chosen for evaluations. The classification accuracies of all the methods are averaged over 10 splits for fair comparison.

*Face Recognition on YaleB-UMIST.* We first use a mixed face database for evaluation. Specifically, we create a mixed face database called YaleB-UMIST that merges the face images of extended YaleB [22] and UMIST [23]. Since YaleB dataset has 38 individuals and around 64 near frontal images under different illuminations per individual, and UMIST database has 575 images of 20 persons with mixed race/gender/ appearance, the mixed YaleB-UMIST face dataset will be more challenging for face recognition. We have exhibited several image examples of YaleB-UMIST in the left of Fig. 6. In this study, we corrupt

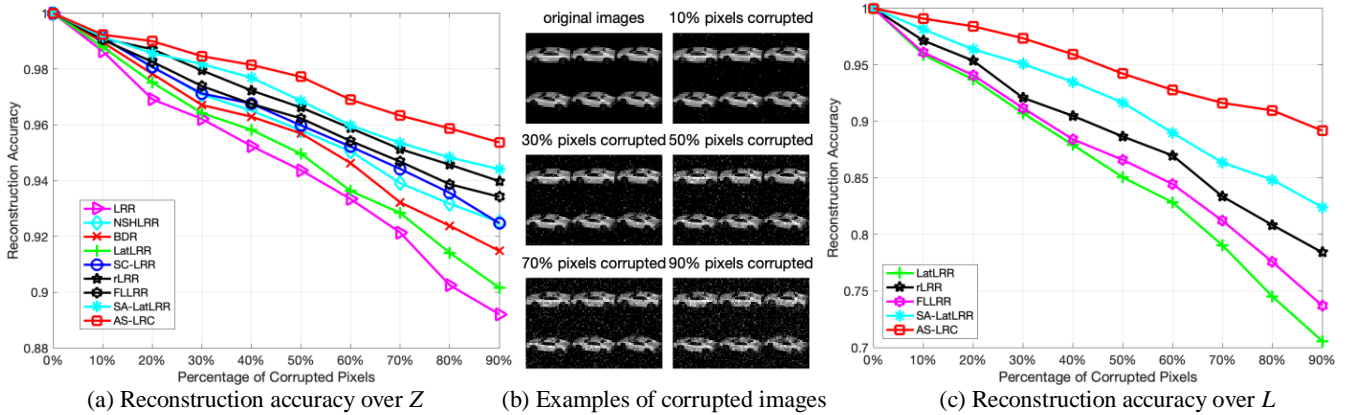

**Fig. 4:** Quantitative image de-noising evaluation results based on the object images of COIL-20 with random pixel corruptions.

(a) Reconstruction accuracy over $Z$    (b) Examples of corrupted images    (c) Reconstruction accuracy over $L$

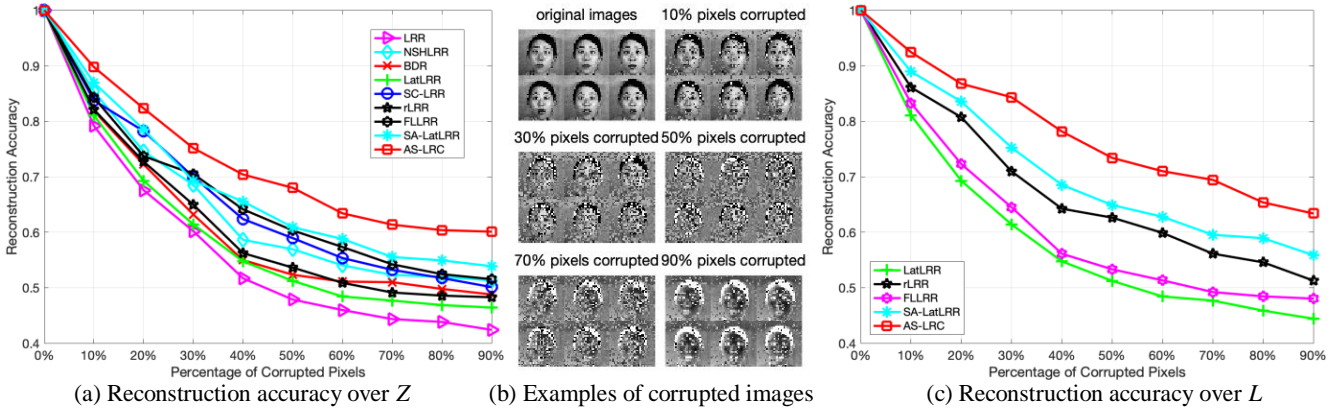

(a) Reconstruction accuracy over $Z$    (b) Examples of corrupted images    (c) Reconstruction accuracy over $L$

**Fig. 5:** Quantitative image de-noising evaluation results based on the face images of JAFFE with inverted pixel value corruptions.

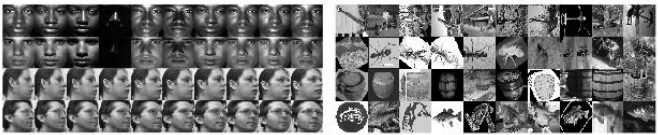

**Fig. 6:** Image examples of the mixed YaleB-UMIST (left) and Caltech 101 databases (right).

40% of the training image pixels by adding random Gaussian noise with zero mean and variance 250. The number of training images per person is set to 10, 12, 14 and 16, respectively. The average result (%) with standard deviation (STD) and the best result are shown in Table 2. We can find from the results that the performance of each method can be enhanced by increasing the training numbers. AS-LRC outperforms other algorithms by delivering comparable and even higher accuracies in most cases. FLLRR, I-LSPFC, rLRR and SA-LatLRR also work well. PCA is the worst one due to the intractable noise and plain model.

***Object Recognition on Caltech 101 [24].*** Caltech 101 has 9111 object images of 101 different categories, and each object category contains between 40 and 800 images on average. In this study, we extract 40 images per subject from 56 categories to construct a subset. Some image examples are shown in Fig. 6. The percentage of corrupted pixels is fixed to 50%, and vary the number of training images per subject from 5 to 20 with step 5. We show the averaged recognition results in Table 3, from which we see that our method still delivers higher accuracies than its competitors in most cases, especially when the number of training samples is relatively small.

***Object Recognition on ETH80 [24].*** In this section, we test each method for recognizing the object images of ETH80 which contains totally 3280 images from 8 big categories, and each big category contains 10 small sub-categories. In this study, we select 41 images from each of the eight big categories to form a subset and reduce the dimensionality to 800 by using PCA. The percentage of corrupted pixels is fixed to 50%, and vary the number of training images per subject from 10 to 22 with step 4. The results are shown in Table 4, from which we can find that our method outperforms than other methods in most cases.

**Table 2**: Face recognition performance vs. different training numbers on mixed YaleB-UMIST face database.

| Setting Methods | YaleB-UMIST(10 train) Mean±STD | Best | YaleB-UMIST(12 train) Mean±STD | Best | YaleB-UMIST(14 train) Mean±STD | Best | YaleB-UMIST(16 train) Mean±STD | Best |
|---|---|---|---|---|---|---|---|---|
| PCA | 54.36±3.21 | 56.25 | 59.64±4.27 | 61.73 | 62.74±2.61 | 65.05 | 67.47±2.31 | 68.94 |
| PCA-L1 | 62.62±1.35 | 65.36 | 66.62±1.26 | 67.74 | 71.32±2.32 | 72.56 | 75.72±4.36 | 77.37 |
| IRPCA | 65.36±1.62 | 66.18 | 69.23±2.62 | 70.96 | 72.68±1.86 | 73.48 | 76.43±3.27 | 79.08 |
| rLRR | 69.56±2.36 | 70.38 | 72.84±3.47 | 74.46 | 76.43±2.41 | 77.91 | 79.96±2.71 | 80.15 |
| LatLRR | 64.67±4.32 | 67.32 | 67.42±2.34 | 68.81 | 72.64±1.84 | 73.06 | 75.85±1.98 | 77.83 |
| SA-LatLRR | 67.32±2.42 | 68.92 | 72.34±2.14 | 75.36 | 75.45±2.61 | 78.49 | 78.51±3.63 | 81.03 |
| FLLRR | 66.34±2.42 | 68.92 | 68.92 | 68.92 | 73.53 | 74.64±2.26 | 76.17 | 77.42±2.38 | 80.12 |
| I-LSPFC | 71.26±2.93 | 73.72 | 75.74±2.18 | 78.47 | 78.24±2.35 | 82.94 | 80.42±3.74 | 82.63 |
| AS-LRC | **75.57±2.35** | **79.41** | **77.53±1.93** | **79.41** | **79.23±1.74** | **84.72** | **81.74±2.17** | **85.42** |

Table 3: Object recognition performance vs. different training numbers on Caltech 101 object database.

| Setting Methods | Caltech 101(5 train) Mean±STD | Best | Caltech 101(10 train) Mean±STD | Best | Caltech 101(15 train) Mean±STD | Best | Caltech 101(20 train) Mean±STD | Best |
|---|---|---|---|---|---|---|---|---|
| PCA | 41.59±2.51 | 44.72 | 47.45±4.67 | 51.73 | 52.48±3.35 | 55.83 | 55.97±3.46 | 57.84 |
| PCA-L1 | 46.81±1.46 | 47.92 | 51.80±2.85 | 53.70 | 57.03±4.64 | 59.83 | 62.72±4.42 | 64.09 |
| IRPCA | 46.84±2.42 | 49.59 | 53.68±3.65 | 56.91 | 57.59±3.48 | 58.95 | 64.62±3.74 | 66.52 |
| rLRR | 47.07±3.31 | 49.03 | 56.04±3.63 | 57.97 | 62.78±2.93 | 64.38 | 69.96±2.46 | 72.43 |
| LatLRR | 45.36±3.42 | 49.81 | 54.83±4.48 | 56.85 | 60.36±3.64 | 63.81 | 65.85±4.57 | 59.81 |
| SA-LatLRR | 48.90±2.51 | 53.15 | 55.36±3.26 | 57.42 | 63.98±2.39 | 65.83 | 69.83±3.68 | 74.38 |
| FLLRR | 47.94±2.53 | 51.35 | 54.38±3.12 | 56.68 | 63.46±1.94 | 64.84 | 68.31±2.24 | 71.56 |
| I-LSPFC | 49.34±3.73 | 53.95 | 56.95±2.47 | 58.64 | 64.84±3.64 | 66.93 | 70.36±3.56 | 74.73 |
| **AS-LRC** | **53.52±3.35** | **56.89** | **58.39±3.36** | **60.43** | **65.83±2.74** | **68.62** | **72.65±3.16** | **76.48** |

Table 4: Object recognition performance vs. different training numbers on ETH80 object database.

| Setting Methods | ETH80 (10 train) Mean±STD | Best | ETH80 (14 train) Mean±STD | Best | ETH80 (18 train) Mean±STD | Best | ETH80 (22 train) Mean±STD | Best |
|---|---|---|---|---|---|---|---|---|
| PCA | 45.76±1.84 | 46.26 | 49.37±2.11 | 52.31 | 56.36±1.64 | 57.72 | 60.43±1.53 | 62.97 |
| PCA-L1 | 47.58±2.17 | 48.86 | 51.73±1.96 | 53.73 | 58.43±1.86 | 59.84 | 62.35±1.86 | 63.86 |
| IRPCA | 48.34±1.68 | 49.75 | 52.36±1.34 | 53.47 | 59.74±1.76 | 60.38 | 62.97±1.67 | 63.87 |
| rLRR | 51.37±1.57 | 52.37 | 57.73±1.86 | 58.74 | 63.48±1.17 | 64.32 | 67.42±1.36 | 69.32 |
| LatLRR | 47.52±2.37 | 49.43 | 52.49±1.85 | 53.86 | 60.12±1.27 | 62.48 | 62.89±2.27 | 64.23 |
| SA-LatLRR | 52.86±1.64 | 53.87 | 60.13±1.87 | 62.74 | 66.84±1.93 | 68.76 | 70.89±1.76 | 72.98 |
| FLLRR | 49.54±1.43 | 51.38 | 53.86±1.87 | 55.89 | 62.85±1.84 | 63.32 | 64.74±1.86 | 65.83 |
| I-LSPFC | 52.64±1.37 | 53.86 | 59.85±2.57 | 62.26 | 65.38±1.58 | 66.46 | 70.42±1.87 | 72.90 |
| **AS-LRC** | **56.61±2.16** | **57.73** | **63.34±2.53** | **65.32** | **69.64±1.28** | **71.31** | **73.23±2.26** | **75.28** |

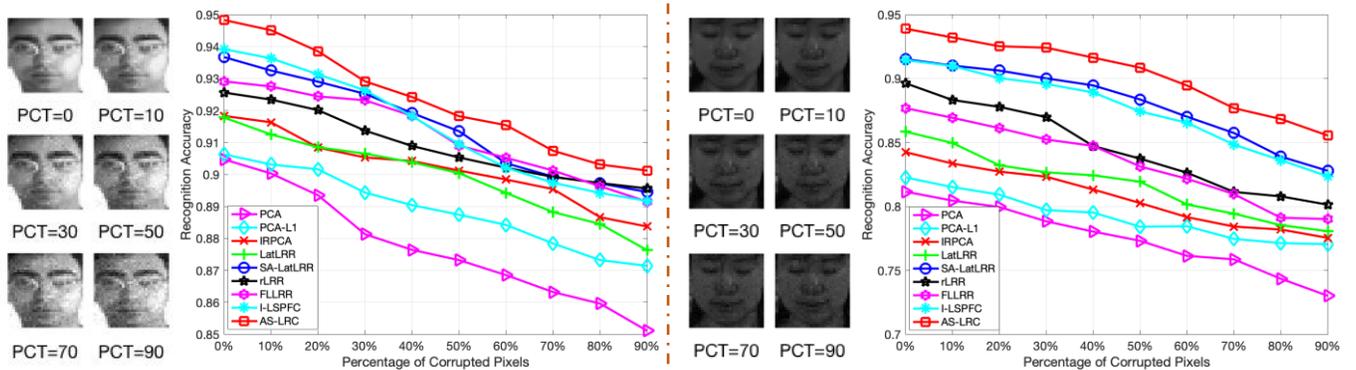

**Fig. 7:** Face recognition performance of each algorithm with varying levels of corruption on AR (left) and CMU PIE (right) databases.

*D. Noisy Face Recognition vs. Different Pixel Corruptions*

We evaluate our algorithm for recognizing face images under fixed training number and different levels of pixel corruptions. AR face [20] and CMU PIE face databases [21] are evaluated. The accuracy rates are averaged based on 10 random splits.

*Recognition Results on AR.* The face images of AR feature frontal view faces with different expressions, illuminations, and occlusions (sun glasses and scarf). In this simulation, we set the number of training images per person is set to 10 and vary the percentage (PCT) of corrupted pixels. To corrupt the data, white Gaussian noise with SNR=20 added into the training set. The accuracies are shown in Fig. 7. We can find that the increasing percentage of corruptions can degrease the performance of each model. But our method delivers higher accuracies than others in most cases. SA-LatLRR and I-LSPFC also perform well, and PCA is the worst one due to its inability to handle noise in data.

*Recognition Results on CMU PIE.* This database has images captured under varying pose, illumination and expression. In this study, we extract a subset by selecting 36 images from each individual (totally 2448 face images) with different poses and lighting condition. We add white Gaussian noise with SNR=15 to corrupt data and choose 10 images per individual for training. The average results are exhibited in Fig. 7. We find that our method performs better than the other competitors in most cases due to the enhanced robustness. SA-LatLRR and I-LSPFC gain comparable results, and both are superior to other methods.

*E. Convergence Analysis*

We provide some numerical results to show the convergence behavior of our AS-LRC. In this study, the Extended YaleB, MIT CBCL, CMU PIE and AR databases are applied. For each set, we choose the fixed number of labeled images (twenty for MIT CBCL, ten for CMU PIE, eight for AR and ten for YaleB) to obtain the averaged results. We mainly compared AS-LRC with the most related algorithms i.e., LatLRR and SC-LRR.

The results averaged with 10 times iterations are shown in Fig. 8, where horizontal axis in each subfigure is the number of iterations and the vertical axis is convergence error produced in iterative process. From the numerical results, we clearly see that the convergence error of our method can converge closely to zero. More importantly, we observe that the convergence speed of our algorithm is ranging from 50 to 150 iterations, which is faster than those of LatLRR and SC-LRC.

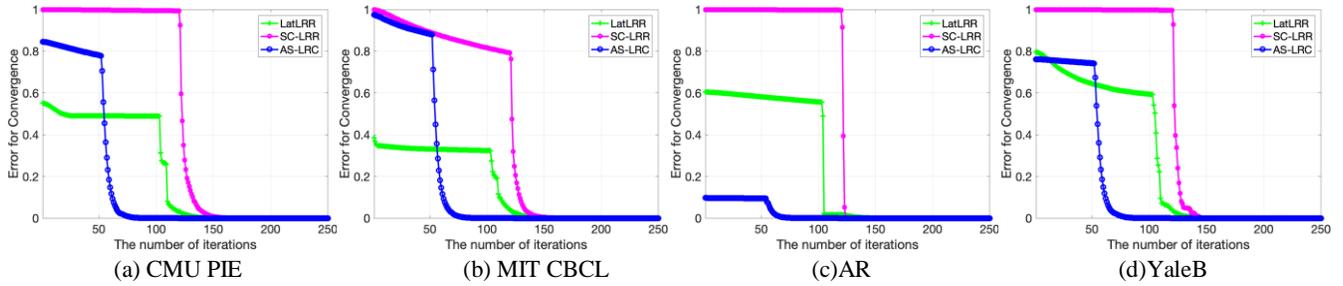

**Fig. 8:** The convergence results of LatLRR, SC-LRR and our AS-LRC based on the CMU PIE, MIT CBCL, AR and YaleB databases.

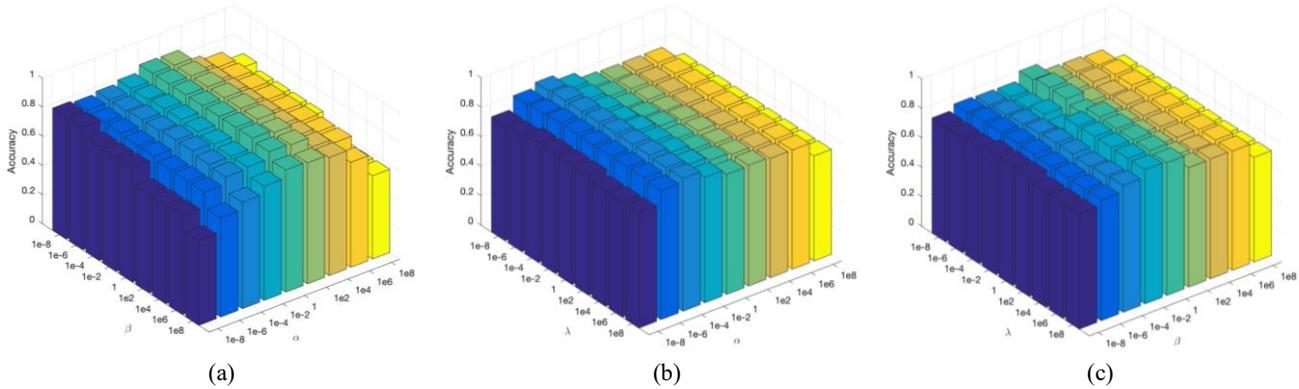

**Fig. 9:** Parameter sensitivity analysis under different parameters on AR: (a) effects of tuning $\alpha$ and $\beta$ on the results by fixing $\lambda = 0.015$, (b) effects of tuning $\alpha$ and $\lambda$ on the results by fixing $\beta = 0.1$, and (c) effects of tuning $\beta$ and $\lambda$ on the results by fixing $\alpha = 0.01$.

*F. Parameter Selection Analysis*

In this study, we mainly analyze the parameter sensitivity of our proposed AS-LRC algorithm. Since parameter selection issue still remains an open issue, so a heuristic way is to select the most important ones. The proposed AS-LRC has three model parameters, i.e., $\alpha$, $\beta$ and $\lambda$, so we can fix one of the parameters and explore the effects of other two on the performance by grid search from the candidate set that ranges from $\{10^{-8}, 10^{-6}, \ldots, 10^{6}, 10^{8}\}$. In this section, the AR face database is employed. We randomly select 8 face images of each individual from the AR database for training and test on the rest. The image recognition accuracy obtained from the test set is employed as the evaluation metric. For each pair of parameters, we average the results based on ten random splits of training and testing samples with varied parameters from the candidate set.

The parameter selection results are shown in Fig. 9, where three groups of results are presented. We can find that our AS-LRC can perform well in a wide range of parameters for each database, since it obtains promising results under the most of parameter settings. That is, our AS-LRC is not sensitive to the model parameters by delivering robust performance. To show the best results of our algorithm, we choose the most important parameters for various datasets, and determine the parameters in a similar way, which is due to the fact that various datasets tends to deliver different distributions and structures.

## V. CONCLUSION AND FUTURE WORK

In this paper, we proposed an Adaptive Structure-constrained Low-Rank Coding (AS-LRC) framework for recovering low-rank and sparse subspaces in a latent learning manner. The new model seamlessly integrates an adaptive auto-weighting based block-diagonal structure-constrained low-rank representation and the group sparse salient feature extraction. By computing an auto-weighting matrix based on locality-adaptive features and multiplying it by the low-rank codes for minimization, one can clearly preserve the locality structures of salient features, force the codes to be block-diagonal and avoid the tricky issue when computing the neighborhood of samples.

We have evaluated the performance of our method on image reconstruction, image de-noising and image recognition. The numerical results demonstrated the remarkable representation and recognition abilities of our method compared with several closely-related models. In future work, we will discuss how to extend our algorithm to semi-supervised representation scenario [29-30][36] using both labeled and unlabeled data. The optimal selection issue of model parameters will also be explored.

ACKNOWLEDGMENT

This work is partially supported by the National Natural Science Foundation of China (61672365, 61732008, 61725203, 61622305, 61871444, 61572339), and Fundamental Research Funds for the Central Universities of China (JZ2019HGPA01-02). Dr. Zhao Zhang is the corresponding author.